\documentclass[11pt,a4paper]{article}
\usepackage[hyperref]{acl2021}

\usepackage{times}
\usepackage{latexsym}

\usepackage{microtype}

\aclfinalcopy 
\def\aclpaperid{3462} 

\usepackage{xspace}
\usepackage{amsmath}
\usepackage{amssymb}
\usepackage{mathtools}
\usepackage{enumitem}
\usepackage{multirow,bigdelim}
\usepackage{comment}
\usepackage{framed}
\usepackage{booktabs}
\usepackage{color, colortbl}
\usepackage{tabularx}
\usepackage{soul}
\usepackage{xcolor}
\usepackage{arydshln} 

\newcommand\T{\rule{0pt}{2.4ex}}

\newcommand{\sysname}{\textsc{ReadOnce} Transformers\xspace}
\newcommand{\sysnamesingular}{\textsc{ReadOnce} Transformer\xspace}
\newcommand{\sysshortname}{\textsc{ReadOnce}\xspace}

\newcommand{\sysenc}{\textsc{ReadOnce} Representations\xspace}

\definecolor{DocEnc}{rgb}{0.85, 0.9, 0.99}
\definecolor{RepText}{rgb}{0.88, 0.83, 0.9}
\newcommand{\CtxtEncNoc}{Document Encoder\xspace}
\newcommand{\RepTextNoc}{Representation+Text Model\xspace}
\newcommand{\CtxtEnc}{{\sethlcolor{DocEnc}\hl{Document Encoder}}\xspace}
\newcommand{\RepText}{{\sethlcolor{RepText}\hl{Representation+Text Model}}\xspace}

\newcommand{\hotpot}{HotpotQA\xspace}

\newcommand{\squad}{SQuAD\xspace}
\newcommand{\unsup}{UQA\xspace}
\newcommand{\nqa}{Narr.QA\xspace}
\newcommand{\xsum}{XSUM\xspace}

\newcommand{\bart}{BART\xspace}

\newcommand{\doc}{\ensuremath{d}}
\newcommand{\tok}{\ensuremath{t}}
\newcommand{\enco}{\ensuremath{h}}
\newcommand{\rotvec}{\ensuremath{r}}

\title{\sysname:\\ Reusable Representations of
 Text for Transformers}

\author{
Shih-Ting Lin\thanks{\ \ The author's work was primarily done during an internship at the Allen Institute for AI.} \;\;\;\; Ashish Sabharwal$^\dagger$ \;\;\;\; Tushar Khot$^\dagger$ \\ \\
 $^\ast$ University of Texas, Austin, U.S.A.\\
 $^\dagger$ Allen Institute for AI, Seattle, U.S.A. \\
 {\small \tt j0717lin@utexas.edu, \{ashishs,tushark\}@allenai.org}
}

\date{}

\begin{document}

\maketitle

\begin{abstract}
We present \sysname, an approach to convert a transformer-based model into one that can build an information-capturing, task-independent, and compressed representation of text. The resulting representation is reusable across different examples and tasks, thereby requiring a document shared across many examples or tasks to only be \emph{read once}. This leads to faster training and evaluation of models. Additionally, we extend standard text-to-text transformer models to Representation+Text-to-text models, and evaluate on multiple downstream tasks: multi-hop QA, abstractive QA, and long-document summarization. Our one-time computed representation results in a 2x-5x speedup compared to standard text-to-text models, while the compression also allows existing language models to handle longer documents without the need for designing new pre-trained models.\footnote{For model code, see \url{https://github.com/allenai/readonce}}

\end{abstract}

\section{Introduction}

Transformer-based large scale language models (LMs)~\cite{radford2018improving,devlin-etal-2019-bert}
are task-independent models that are surprisingly effective when directly fine-tuned on many different end-tasks~\cite{squad,wang2018glue,superglue}. 
However, this approach relies heavily on using end-task supervision to learn to solve two sub-problems simultaneously: extract information\footnote{By ``extract information", we mean implicitly or explicitly compute some representation of the document.} from an input document $D$ and solve the end-task (e.g., answer a question about $D$). This incentivizes LM-based models to learn to extract only task-specific---and even example-specific---information when fine-tuned on the end-task. For example, a Question Answering (QA) model may learn to only extract the answer from $D$ given the input question. 

\begin{figure}[t]
    \centering
    \includegraphics[width=0.76\columnwidth]{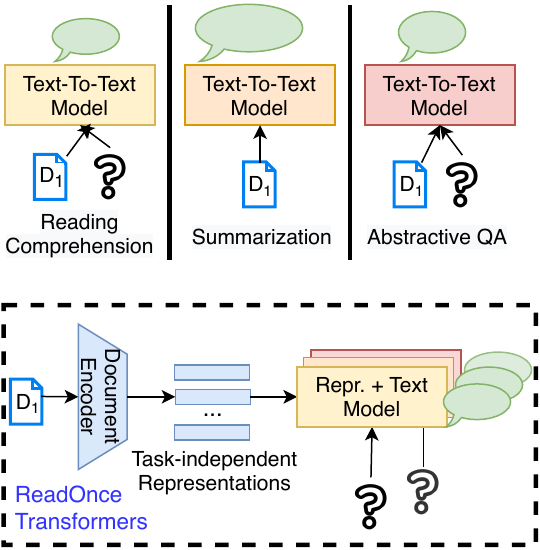}
    \caption{
        \sysname: Rather than learning to extract information specific to each end-task, we use transformer-based encoders to build task-independent, reusable document representations once, and feed them into various representation+text transformer models trained for end-tasks.
    }
    \label{fig:intro_fig}
    \vspace{-2.8ex}
\end{figure}

This strategy, while effective on many datasets, is also inefficient. First, it requires model's pre-trained weights to be fine-tuned separately  for each end-task, even though the sub-problem of gathering the information content of the input document $D$ is shared across tasks. Second, each $D$ must be re-read from scratch in the context of each example (e.g., once for each question) even when many examples share $D$. Not only is this computational redundancy undesirable, slow inference can quickly become a bottleneck in deployed, real-time systems if models with billions of parameters must re-read $D$ for every input query. 

Inspired by humans' ability to read a document and extract key information from it without having to know the use case in advance, we ask the following question: Can we use transformer-based LMs to \emph{build compressed representations of text} that are example- and task-independent, and hence reusable? Further, can we extend text-to-text transformer architectures to consume such representations in conjunction with text?

Prior representation learning approaches attempt to capture the meaning of sentences into a continuous vector~\cite{infersent,skipthought,sentence-bert}. While they have been effective on downstream classification tasks, it is unclear whether they can capture the information content of entire paragraphs. Moreover, these approaches focus on building fixed-length representations that are used as the input features for task-specific classifiers. In contrast, our goal is to (a) use transformer-based LMs to build compressed representations that \emph{scale with the document size}, and (b) \emph{combine them with example-specific text inputs} to produce the more general text output.

To this end, we propose an approach to convert any encoder-decoder based transformer LM (such as BART~\cite{lewis2019bart}) into a new architecture termed \sysnamesingular, with two key parts: (1) a \CtxtEnc that reads documents only once to create compressed, information-capturing, reusable representations that we refer to as \sysenc (2) a \RepText that consumes these document representations together with task- and example-specific plain text (e.g., a question) to produce text output (e.g. an answer). To ensure that our compressed representations capture the key facts, we use supervision from two factoid QA datasets, \squad~\cite{squad} and UnsupervisedQA~\cite{unsupervisedqa} to train \sysname. To solve an end-task, we only need to compute the \sysenc of the documents once and only train the \RepTextNoc to perform the end-task.

Our experiments demonstrate that these representations are more effective at capturing information compared to baseline approaches. Our representations also generalize to other tasks such as multi-hop QA~\cite{hotpotqa}, abstractive QA~\cite{Kocisk2018NarrativeQA}, and summarization~\cite{Narayan2018DontGM}. Since \sysenc are computed only once, we can train and infer with models 2x-5x faster than standard approaches, with only a marginal drop in accuracy (about 3 F1 points on QA and 4 Rouge-L points on summarization for a 2x speedup). Moreover, the compression ratio parameter $K$ of our representations provides an easy way to trade off computation time with accuracy. Specifically, our analysis suggests that the resulting model has a computation cost of roughly $1/2R + 3/4K^2$ of the base LM, where $R$ is the frequency of document reuse.

Additionally, our compressed representation enables us to efficiently combine information from long (or multiple) documents enabling more accurate long-document summarization~\cite{Cohan2018ADA} without needing costly pre-training of new LMs~\cite{Beltagy2020Longformer,Zaheer2020BigBird}.

\section{Related Work}
Representation learning approaches are commonly used to extract fixed-length sentence embeddings~\cite{infersent,skipthought,wang-etal-2020-cross} from variable-length text inputs. Such fixed length representations have enabled the development of simpler downstream models that do not have to deal with the variable-lengths of textual inputs. However, these representations have mainly been used for simple classification tasks on short input texts~\cite{snli:emnlp2015,wang2018glue}. The word-level representations from RNNs or transformers are also variable-length, but uncompressed. While such representations have been re-used with RNNs~\cite{elmo} and are easy to combine with text input, it is not immediately clear how to combine representations from transformers with text, which is what we propose.

Recent work~\cite{sentence-bert,quase,Artetxe2019MassivelyMS,dpr} has tried building document-embedding using large-scale language models as well. However these fixed-length representations have mostly been built to identify similar documents~\cite{sentence-bert,dpr} and are not used directly for QA. QuASE~\cite{quase}, also used question-answering supervision for transfer learning but do not produce re-usable representations. \citet{Artetxe2019MassivelyMS} learned multi-lingual sentence embeddings that may be able to capture the knowledge present in a sentence but they were designed for BiLSTMs. Some large-scale LMs have been especially designed to handle long documents~\cite{yang2019xlnetga,Beltagy2020Longformer,Zaheer2020BigBird} too but need to be pre-trained on large corpora, whereas we can use any pre-trained LM. 

Aspects of our work also bears resemblance to domain adaptation~\cite{Daum2006DomainAF}, transfer learning~\cite{transfer} and multi-task learning~\cite{Caruana1993MultitaskLA} but focuses on learning information-capturing representations from transformer-based models that has not been explored by prior work. While model distillation~\cite{Hinton2015DistillingTK} can also result in speedups, these techniques are orthogonal and can be easily incorporated in our framework (as we show in our experiments).

\section{\sysname}
Our goal in this work is to identify the optimal architecture to extract information-capturing re-usable representations. At the same time, we also need to find the optimal architecture to use such representation in conjunction with text inputs. So at a high level (as shown in Fig.~\ref{fig:intro_fig}), we need to develop two systems: (1) A model to compute the representation, \CtxtEnc\ and (2) A general model for tasks that can consume vector representations and text, \RepText. Given the recent success and generality of encoder-decoder models~\cite{radford2018improving,raffel2020exploring,lewis2019bart}, we focus on developing models for such an architecture. We present the potential choices for each model, with the final model used in our system indicated by a *.

\subsection{\CtxtEncNoc}
\label{subsec:document-encoder}

Given an encoder-decoder model, there are different ways to compute representations for a document $\doc$ with tokens $\{\tok_1, \ldots, \tok_n\}$. We focus on using the output representation generated by the encoder, represented with $\enco_i$ for each token $\tok_i$.

\paragraph{Fixed Length Aggregation.}
The most common approach is to extract a single representation from a sequence of vector~\cite{skipthought,infersent}. While this can be a very compact representation of a document, it tends to be very lossy, especially when dealing with large documents. As a result, these representations are mainly used for classification~\cite{infersent,sentence-bert} or retrieval~\cite{dpr}, and have not been shown to capture the content of the document. E.g, InferSent~\cite{infersent} presented a self-attentive approach to extract sentence embedding using:
\setlength{\abovedisplayskip}{2pt}
\setlength{\belowdisplayskip}{2pt}
\begin{align}
    \rotvec = \sum_i U_{\theta}(\enco_i) \enco_i
     \label{eqn:linear_single}
\end{align}
where $U_{\theta}$ is a function that computes a scalar attention over each $\enco_i$. To reduce information loss, we extend these models to produce $M$ representation vectors by learning $M$ sets of parameters $\theta_j$ for $j \in \{1, \ldots, M\}$, i.e.,
$\rotvec_j = \sum_i U_{\theta_j}(\enco_i) \enco_i$
where $U_{\theta_j}(\enco_i) = e^{\theta_j\enco_i}/\sum_i e^{\theta_j\enco_i}$.

\paragraph{Special Token Representations.}
With the advent of transformer models, another common approach is adding a special \texttt{[CLS]}~\cite{radford2018improving,devlin-etal-2019-bert} or \texttt{<s>}~\cite{liu2019roberta} token to the context. The output representation of this special token can then be used as inputs to classifiers and other down-stream models. Again, a single representation can be lossy, so we generate $M$ representations by inserting multiple special tokens. We can dynamically adjust the number of special tokens based on the input length to produce a variable-length representation. To achieve a compression-ratio of $\frac{1}{k}$, we insert $\frac{N}{k}$ special tokens and use their representations. 

We consider two ways\footnote{More complex designs such as special token embeddings, position embeddings, and indicator features are left as future work.} of inserting special tokens into the context: (1) \textbf{Suffix}: Add them at the end of the context\footnote{Prefixing special tokens generally worsened performance.} (2) \textbf{Interleave}: Add them after every $k$ tokens. While the first approach preserves context continuity, the latter might more directly incentivize the model to capture local context.

\paragraph{Sliding Window Aggregation*.}
We apply the idea of aggregating single-vector representations to generate a variable-length representation. We apply an aggregation function $F$ over sliding windows of size $W$ tokens to capture the local context of the window (akin to CNNs). For a stride length of $S$, this would result in representation vectors:
\begin{align}
    \rotvec_j = F(\{\enco_{S\cdot j}, \cdots, \enco_{S\cdot j + W}\})
\end{align}
where $F \in \{\mu, \alpha, \omega\}$ corresponds to mean-pooling, linear weighting (as described in Eqn.~(\ref{eqn:linear_single})), and max-pooling, respectively. 

\begin{figure}
    \centering
    \includegraphics[width=0.9\columnwidth]{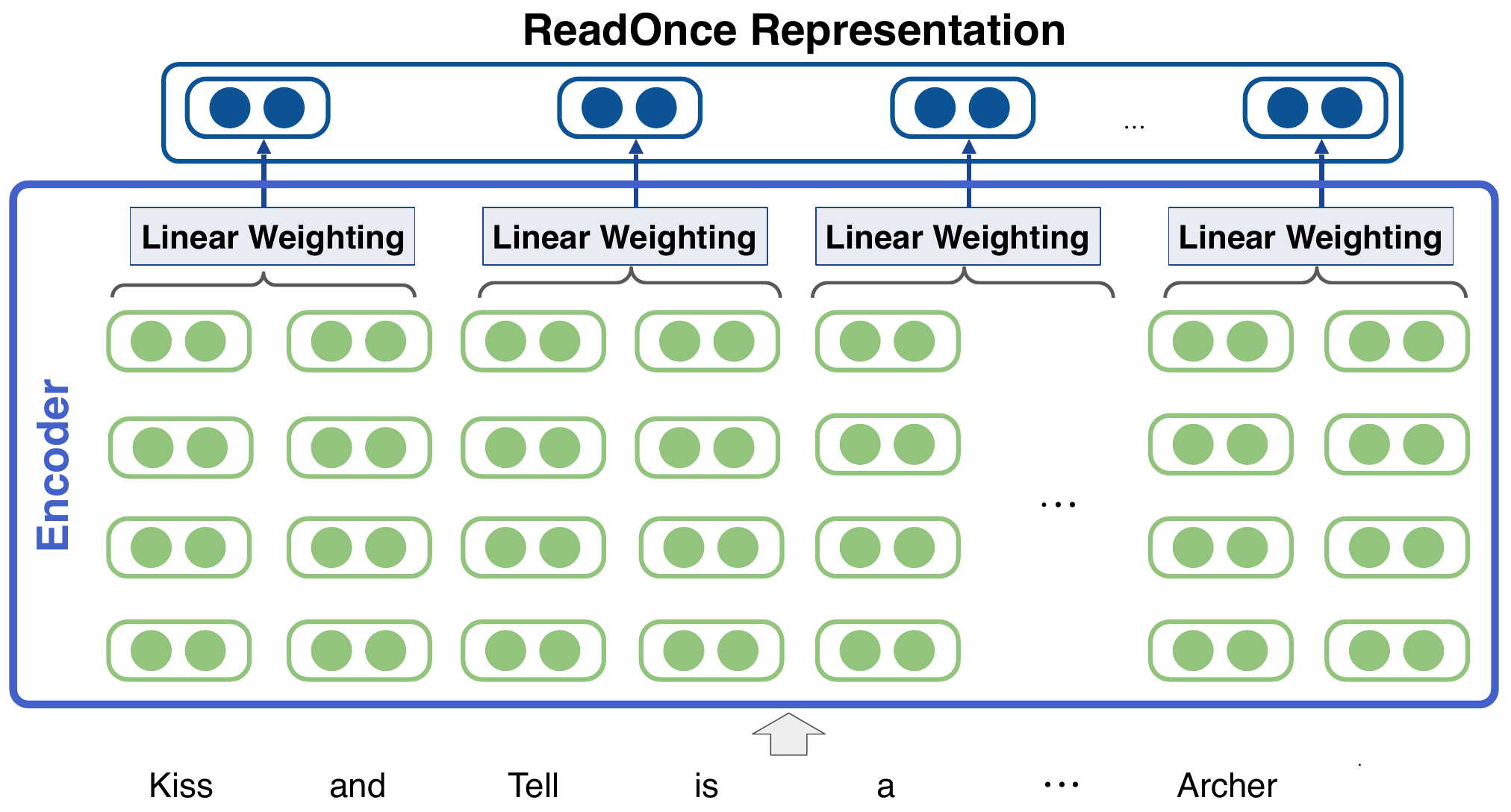}
    \caption{Sliding Window Aggregation approach to extract meaning representations from a transformer-based encoder. Linear weighted sum is used to aggregate each W=2 vectors from the final output layer into a single vector, resulting in the \sysenc with N/2 vectors.}
    \label{fig:encoder}
\end{figure}

Figure~\ref{fig:encoder} shows how we would compute these representations using a window-size of W=2 with no overlap (i.e. S=2) and the linear weighting function. The resulting \sysenc would have $M=N/2$ vectors where N is the number of tokens in the input.

\paragraph{SentenceBERT Baseline.}
For completeness, we also use an existing transformer-based SentenceBert model~\cite{sentence-bert}\footnote{We use the BERT-Large NLI tokens which performed better than the NLI-STSB representations in our experiments} to compute the representation of each sentence in the document. Since the space of these representation might be different, we learn a single-layer feedforward network to project the representations into the right space. For fair comparison to models with variable compression ratio $k$, we also use SentenceBERT representations for a sliding window of $k$ tokens.

\subsection{\RepTextNoc}
\label{subsec:repr+text-model}

Next, we present our modification to downstream task models to use both text and our generated \sysenc. Since most NLP tasks can be re-formulated as a text-to-text problem~\cite{radford2018improving,raffel2020exploring}, we focus on extending text-to-text encoder-decoder models to a (vec+text)-to-text model. 

\begin{figure}
    \centering
    \includegraphics[width=0.9\columnwidth]{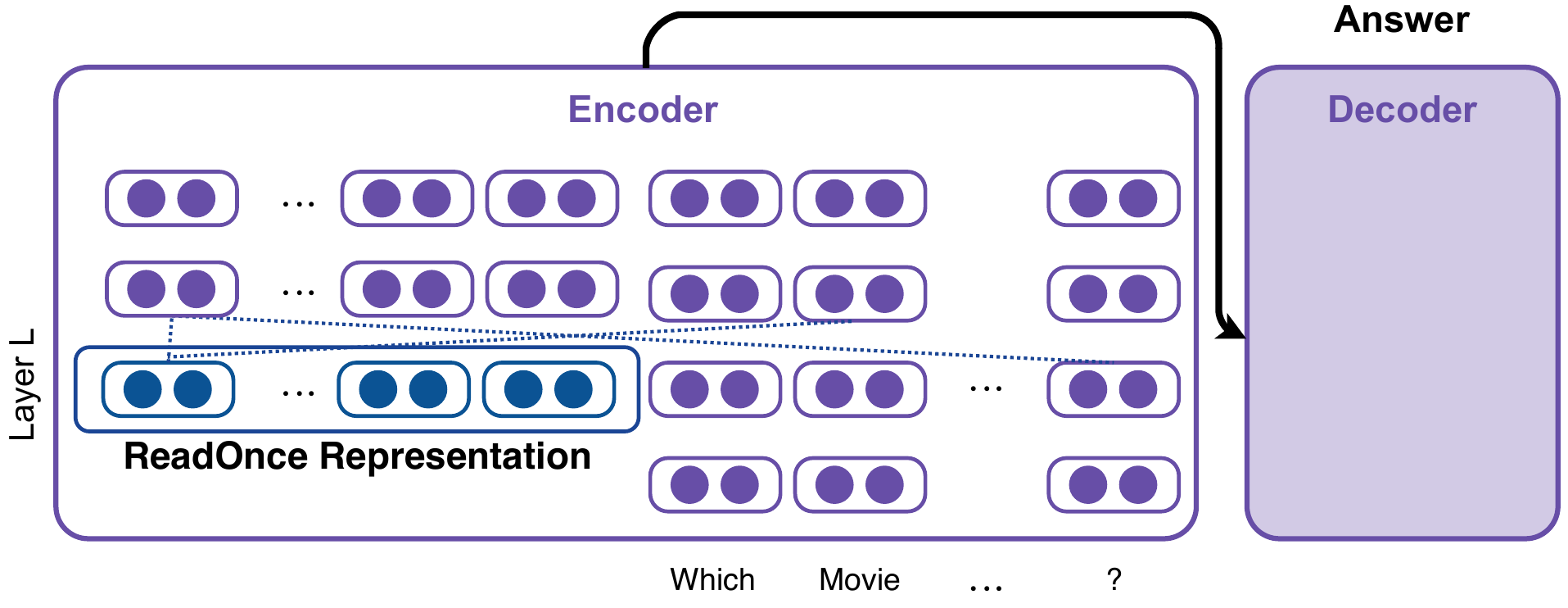}
    \caption{Appending the \sysenc to the $L^{th}$ layer of the encoder to extend standard encoder-decoder models to handle text+vector inputs.}
    \label{fig:reptext}
\end{figure}

\paragraph{Append to Encoder*.}
Since the transformer block in an encoder can handle any input length in each layer, one possible approach is to append the representations to the L$^{th}$ layer of the encoder. This allows the model to focus on parsing the input example text(e.g., question) in the L-1 layers followed by focusing on answering the question in the remaining layers. We show this model in Figure~\ref{fig:reptext} where the encoder only processes the $Q$ tokens of the question for the first L layers. Once the $M$ \sysenc are added to the $L^{th}$ layer, all the subsequent layers produce $M+Q$ vectors by attending over both the representations and text. Finally an unmodified decoder produces the output answer.

\paragraph{Modify Transformer Block Attention.}
Rather than just modifying the input, we consider an alternate approach of modifying the transformer block itself. Similar to PlotMachines~\cite{Rashkin2020PlotMachinesOG}, we view the representation as a memory that the self-attention block can attend over (in addition to the input text). We modify the self-attention blocks in both the encoder and the decoder\footnote{Only modifying the encoder or decoder resulted in slightly lower performance.} to use two separate attention modules for both of these input types and averages the vectors.\footnote{See App.~\ref{app:plotmachines} for the detailed formulas.} With this design, ideally the \RepTextNoc will gain extra capacity to model the interaction between the representation and the input text. 

\subsection{Training \sysshortname via QA}
\begin{figure}
    \centering
    \includegraphics[width=0.9\columnwidth]{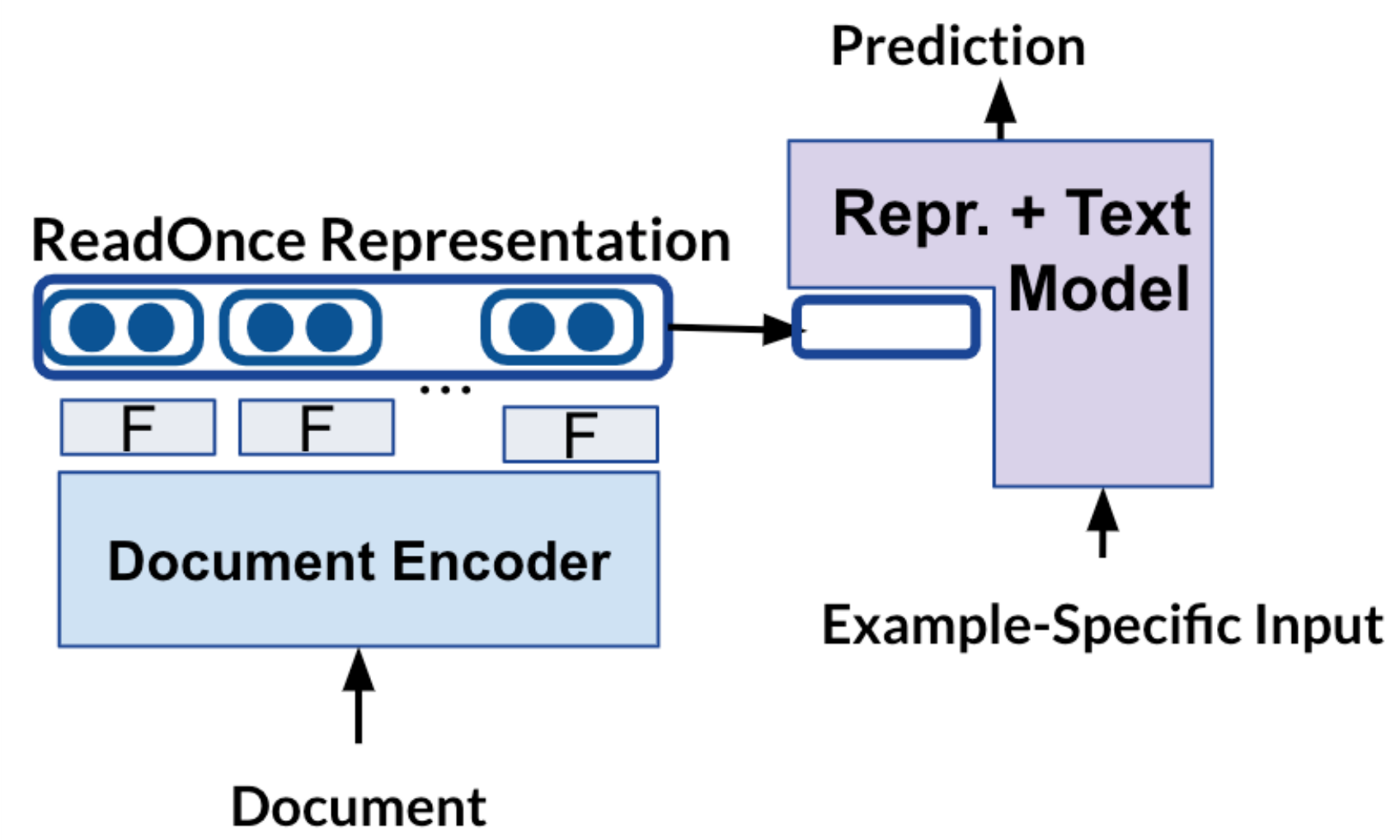}
    \caption{
        \sysname architecture. We use aggregated sliding window representations (Fig.~\ref{fig:encoder}) as \CtxtEnc to compute the \sysenc. We append these representations to the $L^{th}$ layer of the encoder in our \RepText (Fig.~\ref{fig:reptext}). This end-to-end model is fine-tuned on QA tasks to train the \CtxtEnc to extract information-capturing representations. 
    }
    \label{fig:full_arch}
\end{figure}

Given the overall architecture of such a system (shown in Fig.~\ref{fig:full_arch}), we next focus on training this model to produce \sysenc that capture the information present in the document. While prior representation learning models have often focused on classification tasks, we instead use the reading comprehension QA task to ensure this information-capturing property. If a model is able to use just the \sysenc to answer the questions grounded in the document, the representations would contain the information needed to answer such questions.

The key question here is: \emph{Which QA datasets are most suitable for training a compact yet information-capturing document representation?}

Low-level semantic QA datasets~\cite{qamr,qasrl} don't allow for any compression as the questions require the knowledge about every word in the input sentence. More complex multi-hop QA datasets such as \hotpot~\cite{hotpotqa} are also not appropriate, as they focus on learning to reason in addition to capturing the information. Shallow reading comprehension tasks provide a sweet spot between these two extremes, as extracting key information from the given document is sufficient to answer the questions. Further, unlike semantic QA tasks, the questions only focus on the key facts mentioned in a document, which can be captured in a compressed representation. We use two such datasets to train our models: \squad and Unsupervised QA.

\subsection{Downstream Usage of \sysshortname}
To verify the generality of the \sysenc, we train models to perform multi-hop reasoning, abstractive QA and summarization using our learned representations. Specifically, we freeze the \CtxtEnc model and use it to generate the representations for documents. We further fine-tune the \RepText on the downstream task to produce the output label given the \sysenc and any example-specific input.

\section{Representation Learning Experiments}
\label{sec:exp-representation}

We first evaluate the different potential architectural choices for extracting and using document representations discussed in \S\ref{subsec:document-encoder} and \S\ref{subsec:repr+text-model}, respectively. While our main interest is in learning effective representations, we also need to find the optimal \RepTextNoc architecture that can consume the representation. 

\subsection{Training Setup}

We train the entire model on the factoid QA task to ensure that the document representations do capture factual knowledge. We primarily use the \squad reading-comprehension dataset~\cite{squad} containing more than 100,000 crowd-sourced factoid questions. We further augment this dataset with about 500,000 rule-based questions from the UnsupervisedQA (UQA) dataset~\cite{unsupervisedqa}. This increases the size of the training dataset while also introducing question diversity. To avoid these automatically generated questions overwhelming training, we ensure that the same number of questions are selected from both the datasets in each batch (by duplicating \squad questions). In the same vein, we evaluate each model based on their performance on the \squad task.\footnote{The scores on \unsup correlate well with the scores on \squad, with close to 90 F1 for most models.}

Unless otherwise mentioned, we use the \bart-Large model in all our experiments, and optimize the model with cross-entropy loss. We set the learning rate to 1e-5 for the weights initialized from the \bart model, and to 1e-4 for randomly initialized newly added weights, which is shown beneficial in \citet{Peters2019KnowledgeEC}. For other hyper-parameters, we follow \citet{lewis2019bart}. We ran all the experiments on RTX 8000 with 48GB GPU memory. All experiments did not use the complete GPU memory, e.g. experim We kept the batch size and gradient accumulation steps constant (both at 8) across different compression ratios. 

\subsection{Architecture Evaluation}
To be able to evaluate the representations, we need to first select the architecture of the model consuming these representations. 

\subsubsection{\RepTextNoc}
We explore the different choices for the \RepText model discussed in \S\ref{subsec:repr+text-model}, assuming the representation is generated by a simple \CtxtEncNoc model: Mean aggregation over a Sliding Window with both window size and stride being 8 tokens. The results are shown in Table~\ref{tab:reptext}.

\begin{table}[ht]
    \centering
    \footnotesize
    \addtolength{\tabcolsep}{3pt}
    \begin{tabular}{lc|cc}
     \multirow{2}{*}{Architecture} & Design & \multicolumn{2}{c}{\squad} \\
       & Parameters & EM & F1 \\
      \midrule
      Append & L=1\phantom{0} & 35.0 & 52.6 \\
      Append* & L=6\phantom{0} & \textbf{57.4} & \textbf{74.5} \\
      Append & L=12 & 53.5 & 69.2 \\
      ModifyAtt & -- & 55.3 & 71.7 \\
    \end{tabular}
    \caption{Comparing \bart-based architectures for jointly processing continuous representations and text.}
    \label{tab:reptext}
\end{table}

We see that appending \sysshortname representations 
 too early (L=1) or too late (L=12) in the encoder stack is not as effective as appending about half-way (L=6).\footnote{We also experimented with L=3 and L=9, and didn't find any significant gains.} We suspect that appending too early does not allow the model to focus on understanding the question, whereas appending too late does not leave enough room for cross-attention between the question and representations.

Modifying the transformer block to attend over these representations 
 results in a reasonable F1 score on \squad, but it is still outperformed by our simple \emph{Append} architecture. Hence, for the rest of this work, we stick to the simpler architecture of appending the representation at the 6$^{th}$ layer, denoted Append(L=6).

\subsubsection{\CtxtEncNoc}
Given the \RepTextNoc model architecture chosen above, we now explore potential \CtxtEnc architectures to extract \sysenc. For a fair comparison, we ensure that all our evaluated representations use, on average across a dataset, the same number of vectors to represent documents. Table~\ref{tab:docenc} presents EM and F1 scores on \squad for the various architectural choices discussed in \S\ref{subsec:document-encoder}.

\begin{table}[ht]
    \centering
    \footnotesize
    \addtolength{\tabcolsep}{2pt}
    \begin{tabular}{lc|cc}
     \multirow{2}{*}{Architecture} & Design & \multicolumn{2}{c}{\squad} \\
      & Parameters & EM & F1 \\
      \midrule
      SlidingWindow* & W=8,\phantom{0} S=8, F=$\alpha$ & \textbf{58.3} & \textbf{74.7} \\
      SlidingWindow & W=8,\phantom{0} S=8, F=$\mu$ & 57.4 & 74.5 \\
      SlidingWindow & W=8,\phantom{0} S=8, F=$\omega$ & 48.8 & 64.6 \\
      SlidingWindow & W=16, S=8, F=$\alpha$ & 53.1 & 69.6 \\ \cdashline{1-4} 
      \T Suffix & M=N/8 & 44.6 & 58.7\\
      Interleave & M=N/8 & 19.0 & 31.0 \\\cdashline{1-4}
      \T SentenceBERT & M=N/8 & 17.8   & 29.6 \\
      SentenceBERT & \phantom{M}M=\#sent. & 15.2 & 25.2\\\cdashline{1-4}
      \T FixedLength & M=21 & 45.6 & 59.3 \\
    \end{tabular}
    \caption{A comparison of different architectures for extracting continuous representations using \bart encoder. Each approach extracts representations of the same length, namely $1/8^{th}$ of the document length either for each document or on average across the dataset. Since SentenceBERT was trained on sentences, we also show results for SentenceBERT(M=\#sent) with $N/32$ representations per document on average.}
    \label{tab:docenc}
\end{table}

The top 3 rows explore the sliding window architecture 
 with both window size and stride length of 8 (i.e., no overlap between windows), with the three different aggregation functions mentioned earlier. We see that both the mean $\mu$ and the learned weighted sum $\alpha$ have comparable performance on this task, and outperform the max-pooling function $\omega$. We also evaluate the impact of increasing the overlap between windows by increasing the window size (not changing the stride length keeps the average number of vectors constant). For the learned weighted sum function, this results in a 5 point F1 drop, possibly due to the aggregation function having to operate over a larger window.\footnote{We also compared W=8, S=2 with W=2, S=2 in our early experiments and notice a similar trend---the smaller sliding window performs better.}

We next evaluate the approaches inspired by prior work where we add special tokens 
 and use the representations of these tokens. For the \bart model, we use a newly added \texttt{[CLS]} token as our special token. We see from Table~\ref{tab:docenc} that neither appending these tokens at the end nor interleaving them in the input results in representations comparable to the sliding window based approaches.\footnote{Special token prefix scored similar to the Suffix model.} The sliding window representations outperform the pre-trained sentence-based representations from SentenceBERT irrespective of the number of vectors used.\footnote{Even when the SlidingWindow approach is limited to $M=N/32$ vectors, it achieves a higher F1 score (52.4) than SentenceBERT.}  Finally, if we fix the representation length
  to 21 vectors (computed based on the average token length of \squad: 163.7), the learned representations are still not as effective.

\subsection{Final \sysshortname Architecture}
Based on this set of experiments, we use the sliding window architecture for the \CtxtEnc with learned weighted sum as the aggregation function, and append these representations to the 6$^{th}$ layer in the final task-dependent \RepText.

\section{Downstream Task Experiments}
\label{sec:exp-downstream-tasks}

Next, we evaluate the quality of our representations by using them on three downstream tasks, different from the tasks \sysname are trained on, demonstrating faster training and inference. We then show the benefit of using our representation when documents are much longer than the token limit of the underlying LM.

\subsection{Experimental Setup}

\paragraph{Tasks:} We consider three end-tasks, extractive QA, summarization, and abstractive QA, to evaluate our system using the following datasets: (1) \textbf{\hotpot}~\cite{hotpotqa}, a multi-hop reasoning extractive QA dataset. (2) \textbf{XSUM}~\cite{Narayan2018DontGM}, an abstractive News summarization dataset (3) \textbf{NarrativeQA}~\cite{Kocisk2018NarrativeQA}, an abstractive QA dataset where answers are not spans from the input document. More details about these datasets and metrics provided in App.~\ref{app:datasets}


\paragraph{Baselines:} We compare \sysname to \bart-based QA models \emph{that use the document text directly} to answer the given question. Since these models use text directly without any lossy compression, their score is best viewed as an \emph{upper bound} for any representation-based \bart model, including ours. We train the \bart model to generate the answer given the entire document and question (we use ``Summary'' as question for XSUM). In addition to \bart-Large, we evaluate two smaller models: \bart-Base and DistilBART~\cite{distilbart}. Since our representations were trained on \squad and \unsup, we also first fine-tune all our \bart models on the same datasets. 
 

\paragraph{\sysshortname Models:} We freeze the parameters of the \CtxtEnc to generate the representations for all the documents in the datasets. We then use these representations with our \RepText, which is further fine-tuned on each end-task.  To evaluate the impact of our pre-training on QA datasets, we compare our model to the \sysshortname architecture initialized with the \bart model weights, $\sysshortname_\phi$. To illustrate the architecture-independence of our approach and orthogonality to traditional compression methods,  we also train and evaluate \sysshortname models using the BART-Base and DistilBART models. These models were also first trained on \squad+\unsup datasets to learn the document representation. See App.~\ref{appendix:distillation} for more details.

Since our \RepTextNoc can handle a variable number of representation vectors, we can change this \emph{compression ratio}, on-the-fly, without having to change the model architecture. Specifically, we can use a stride-length of $K$ in our \CtxtEncNoc to generate representations that are $1/K^{th}$ of the input length, and then feed them to a downstream model. By reducing  $K$, we can reduce the compression ratio and improve the model accuracy, at the cost of increased runtime.

Interestingly, we discovered that we don't even need to re-train \CtxtEncNoc for each value of $K$. We can achieve a performance comparable to encoders trained individually for each value of $K$, by using the \CtxtEncNoc trained on $K=8$ and only varying $K$ during the fine-tuning step.

\subsection{Representation Quality}
First, we assess the ability of \sysenc to capture document information as compared to using the original document text. As shown in Table~\ref{tab:repr}, our framework at K=2 is about 2x faster than \bart-Large while being only 3 F1 and 4 Rouge-L points behind this model with full access to the text. This demonstrates that \sysenc do capture most of the relevant information in the document. The different compressed models can also result in smaller (DistilBART) or comparable (\bart-Base) speed-ups, but (1) our accuracy vs speed trade-off is more easily controllable via K and (2) we can apply our framework on these models to achieve similar speedups.\footnote{While more recent LMs can outperform \bart (e.g. Pegasus~\cite{pegasus} for summarization), we believe similar tradeoffs can be achieved by applying our framework on these newer models.}

\begin{table}[ht]
    \centering
    \footnotesize
    \renewcommand{\tabcolsep}{1.8mm}
    \begin{tabular}{lccc}
     \multirow{2}{*}{Architecture} & \hotpot & \nqa & \xsum \\
      & \phantom{0}F1 $\mid$ sec.  & R-L $\mid$ sec. & R-L $\mid$ sec. \\
   \midrule
    \bart-Large &&& \\\cmidrule{1-1}
    $\sysshortname_{\phi}$(K=8) & 64.8 $\mid$ 1.0 & 41.9 $\mid$ 1.1 & 32.6 $\mid$ 1.1 \\
    \sysshortname(K=8) & 70.9 $\mid$ 1.0 & 55.7 $\mid$ 1.1 & 31.9 $\mid$ 1.1 \\
    \sysshortname(K=2) & 77.2 $\mid$ 2.0 & 66.7 $\mid$ 2.1 & 35.4 $\mid$ 2.0 \\
     UB$_\text{{\squad+\unsup}}$ & 80.1 $\mid$ 4.0 & 70.5 $\mid$ 5.7 & 37.2 $\mid$ 4.0 \\
     \midrule
     \bart-Base &&& \\\cmidrule{1-1}
    \sysshortname(K=2) & 71.5 $\mid$ 0.6 & 61.3 $\mid$ 0.8 & 31.6 $\mid$ 0.9 \\
    UB$_\text{{\squad+\unsup}}$ & 76.5 $\mid$ 1.3 & 65.8 $\mid$ 1.9 & 32.2 $\mid$ 1.5 \\
    \midrule
    DistilBART &&& \\
     \cmidrule{1-1}
        \sysshortname(K=2) & 75.5 $\mid$ 1.5 & 65.1 $\mid$ 1.8 & 33.4 $\mid$ 1.6 \\
     UB$_\text{{\squad+\unsup}}$ & 80.5 $\mid$ 3.7 & 70.5 $\mid$ 5.3 & 36.5 $\mid$ 3.6 \\
    \end{tabular}
    \caption{
        Performance of \sysname on three datasets, vs.\ corresponding text-to-text transformer models with full access to document text (i.e. \textbf{U}pper \textbf{B}ounds). We also show the training time (secs) per batch for each model. Our representations are able to reduce the training time compared to the upper-bounds at the cost of small drops in accuracy.  
    }
    \label{tab:repr}
\end{table}

Lastly, we note that the $\sysshortname_\phi$ system, which simply uses the \bart model parameters, is about 6 F1 and 14 Rouge-L points behind our model with learned representations. This shows that our model does utilize the factoid questions to learn to extract meaningful
representations --- without our training, the representations obtained from the pre-trained models are not as effective.\footnote{We also observe drops in score when using the \bart model parameters in only the \CtxtEncNoc or only the \RepTextNoc.}

\subsection{Model Efficiency}
One key advantage of \sysenc is that the model needs to \emph{read} the document only \emph{once}, and can reuse pre-computed representations for multiple examples or even multiple tasks. Specifically, if a document is repeated across $R$ examples (the \emph{replication factor}) and we use a compression ratio of $K$, our computation cost per question is roughly only ($1/2R + 3/4K^2$) relative to a baseline seq2seq model (cf.\ App.~\ref{subsec:efficiency} for an analysis). In other words, the higher the replication factor $R$ or the compression ratio $K$, the higher the speedup achieved via \sysenc.

\begin{figure}[ht]
    \centering
    \includegraphics[width=0.485\textwidth]{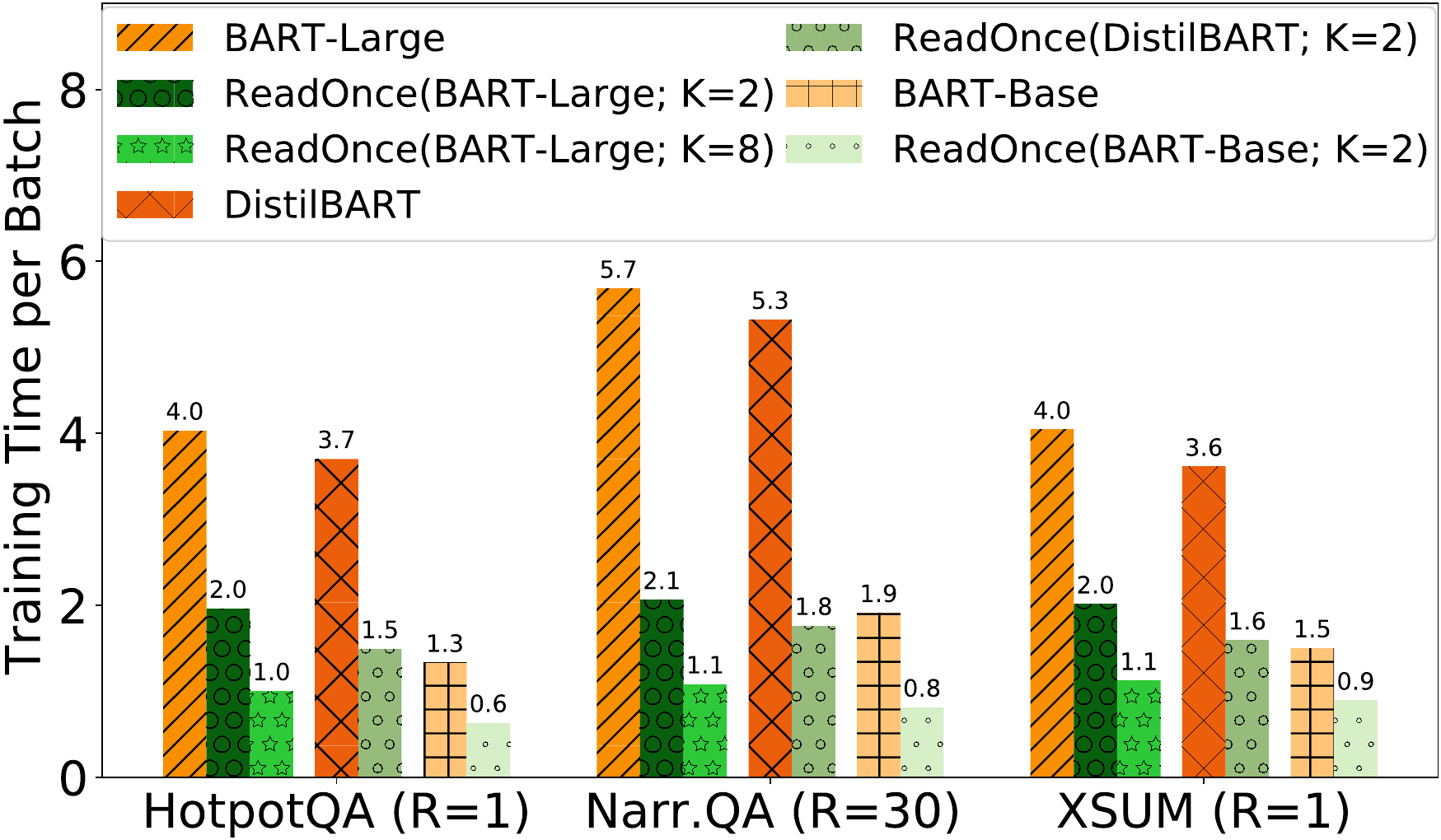}
    \caption{Training time (seconds) per batch. For \sysshortname models, document representations are pre-computed and cached, resulting in a 2-5x speedup.}
    \label{fig:training_time}
\end{figure}

\begin{figure}[ht]
    \centering
    \includegraphics[width=0.485\textwidth]{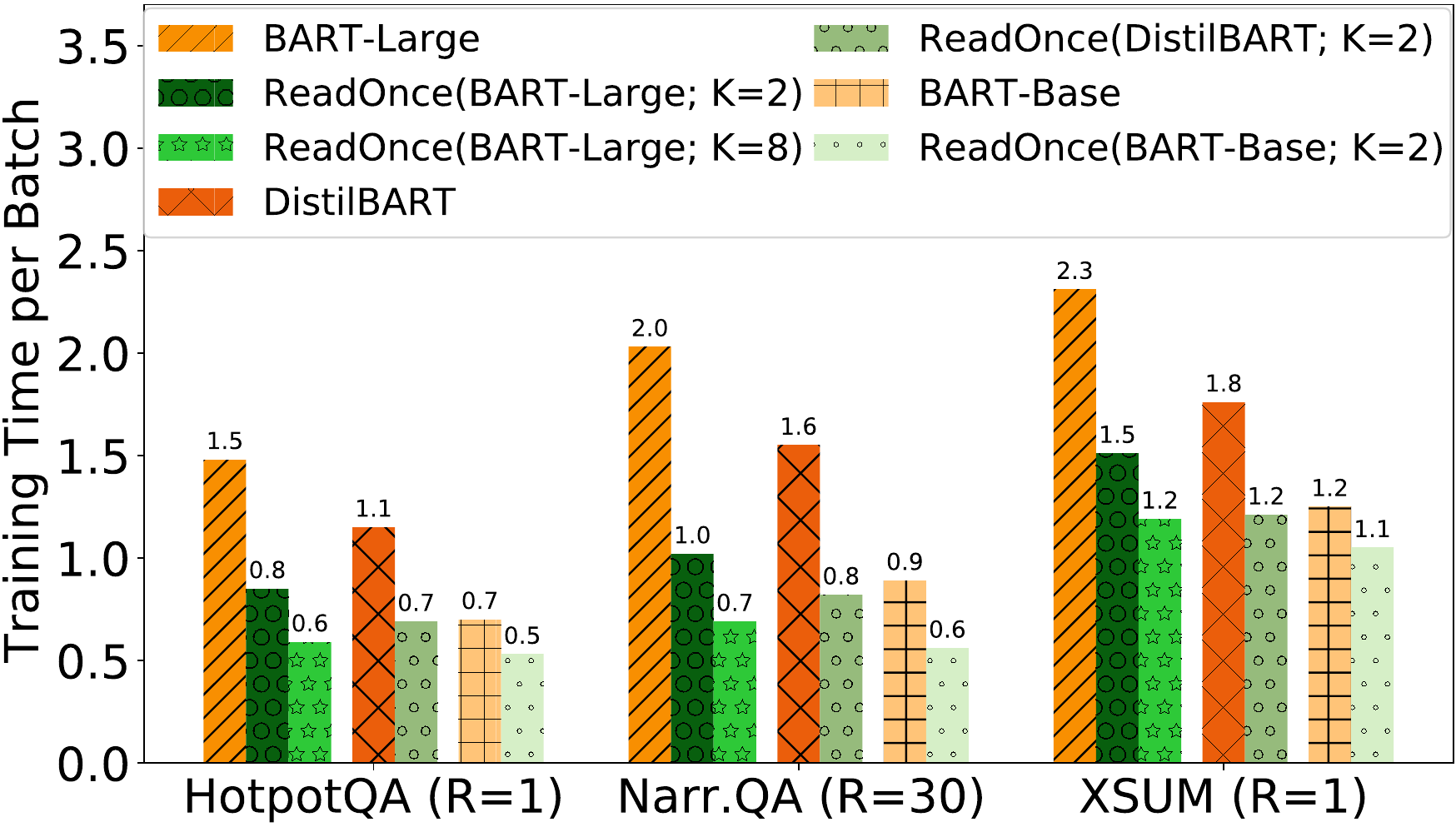}
    \caption{Inference time (seconds) per batch. For \sysshortname models, document representations are pre-computed and cached, resulting in a 2-3x speedup.}
    \label{fig:inference_time}
\end{figure}

Our model exhibits a speedup of 2x-5x in training time compared to the different \bart architectures (Figure~\ref{fig:training_time}). Similarly, we observe a 2x-3x speedup in the inference time (as shown in Figure~\ref{fig:inference_time}), which again plateaus out at K=8. 

Note that the time reported for our model includes the cost of reading \sysenc from disk as well as some fixed costs. These costs form a larger fraction of the overall time for faster models. Hence, while our speedups do not exactly match up to the theoretical analysis, the empirical trends are as expected: we see larger speedups on the NarrativeQA dataset which has a higher replication factor $R$. In general, the R value for our datasets (e.g., R=29.7 for NarrativeQA) is within the range of other datasets (e.g., R=9.4 for NewsQA and R=13.9 for DROP). Note that even when R=1 (e.g., XSUM), we observe a speedup due to the compression ratio K.

\subsection{Efficiency-Accuracy Tradeoff}
\begin{figure}[th]
    \centering
    \includegraphics[width=0.475\textwidth]{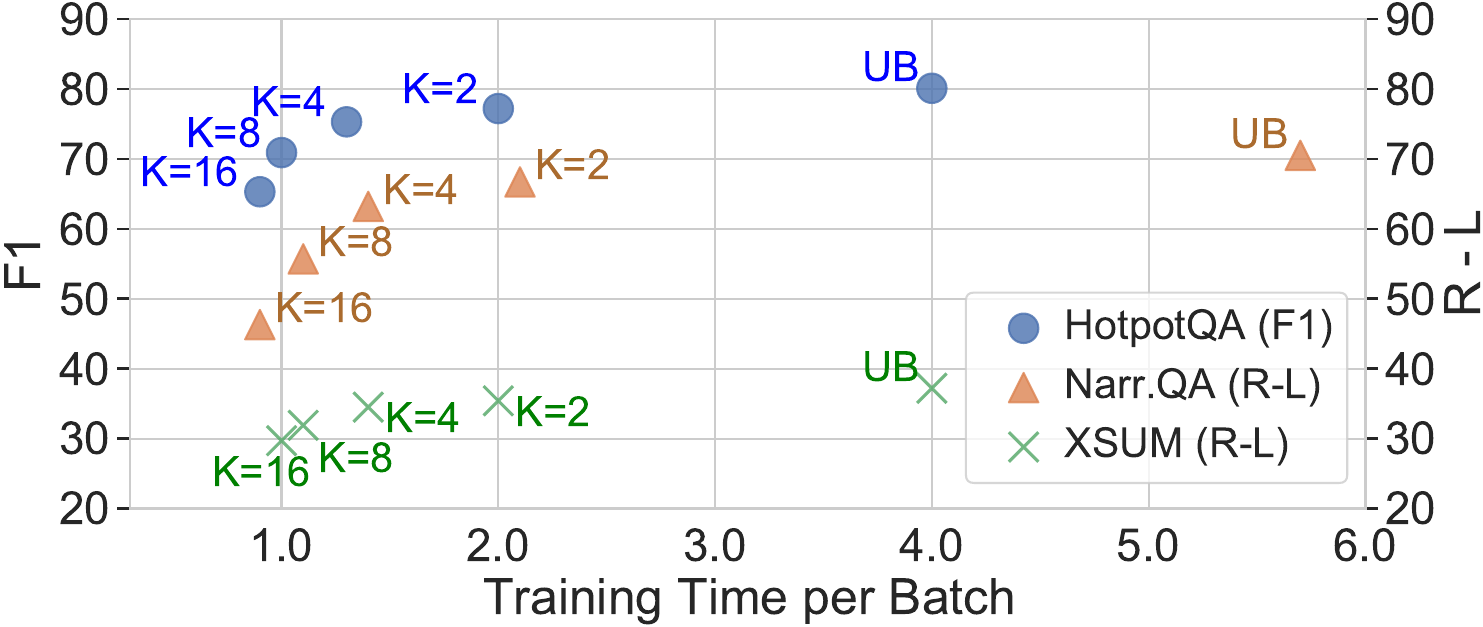}
    \caption{The training time (seconds per batch) vs  performance trade-off achieved by the \sysshortname model with different values of K on our three evaluation tasks. The points annotated with ``K=i"  indicate the \sysshortname models and ``UB" indicate their corresponding \textbf{U}pper \textbf{B}ound. All evaluations use the \bart-Large model. As K increases, the \sysshortname model can be trained faster at the cost of accuracy with diminishing gains after K=4.}
    \label{fig:k_effect}
\end{figure}

We also perform a more fine-grained analysis of the efficiency-accuracy tradeoff in \sysshortname by varying the values of the compression ratio K. As shown in Figure \ref{fig:k_effect}, across all three of our datasets, as the value of K increases, the model's accuracy goes down due to increased compression but so does the training time. As compared to the upper-bound \bart-Large model, we see a large gain in speed when K=2 with diminishing gains as K reaches 8. 

\subsection{Handling Long Documents}
Compressing document representations also enables the downstream model to reason over documents longer than its maximum token length limit T. For example, we can compute representations of document chunks with upto T tokens each and concatenate them together. Since these representations do not rely on any position embeddings in \RepTextNoc, theoretically we can use as many representation vectors as needed.

Given GPU memory limits, lets assume we can only accommodate documents upto length T. Given a compression ratio K, we can compute \sysenc for K such length-T chunks, increasing the capacity of our downstream model to T*K.\footnote{If we allow overlap of O tokens between chunks, the capacity changes to  T*K -- O*(K--1)} For simplicity, we ignore the question as it tends to be much shorter than T.

To assess the impact of increased model capacity, we evaluate our learned representations on the long document summarization task PubMed~\cite{Cohan2018ADA}.\footnote{We also evaluate NarrativeQA, see App.~\ref{appendix:longdoc_nqa}} We follow \citet{Cohan2018ADA} and only include the first 4 sections from each document (average length=2270 tokens). We vary the memory budget from T=512 to T=256 and compare our approach to two \bart seq2seq baselines: a simple truncation baseline with $T/4$ tokens from each section, and a sliding-window baseline often used in QA models for summarization extended here by concatenating summaries from length-T chunks of the input document. For the \sysnamesingular with a compression ratio of K, we can accommodate K*T/4 tokens per section, resulting in a total of T representations from the 4 sections. We choose to obtain these T representations using K/2 chunks from each section, with each chunk containing T/2 tokens.\footnote{See Appendix \ref{appendix:pubmed-setup} for more details.}

\begin{figure}[ht]
    \centering
    \includegraphics[width=0.485\textwidth]{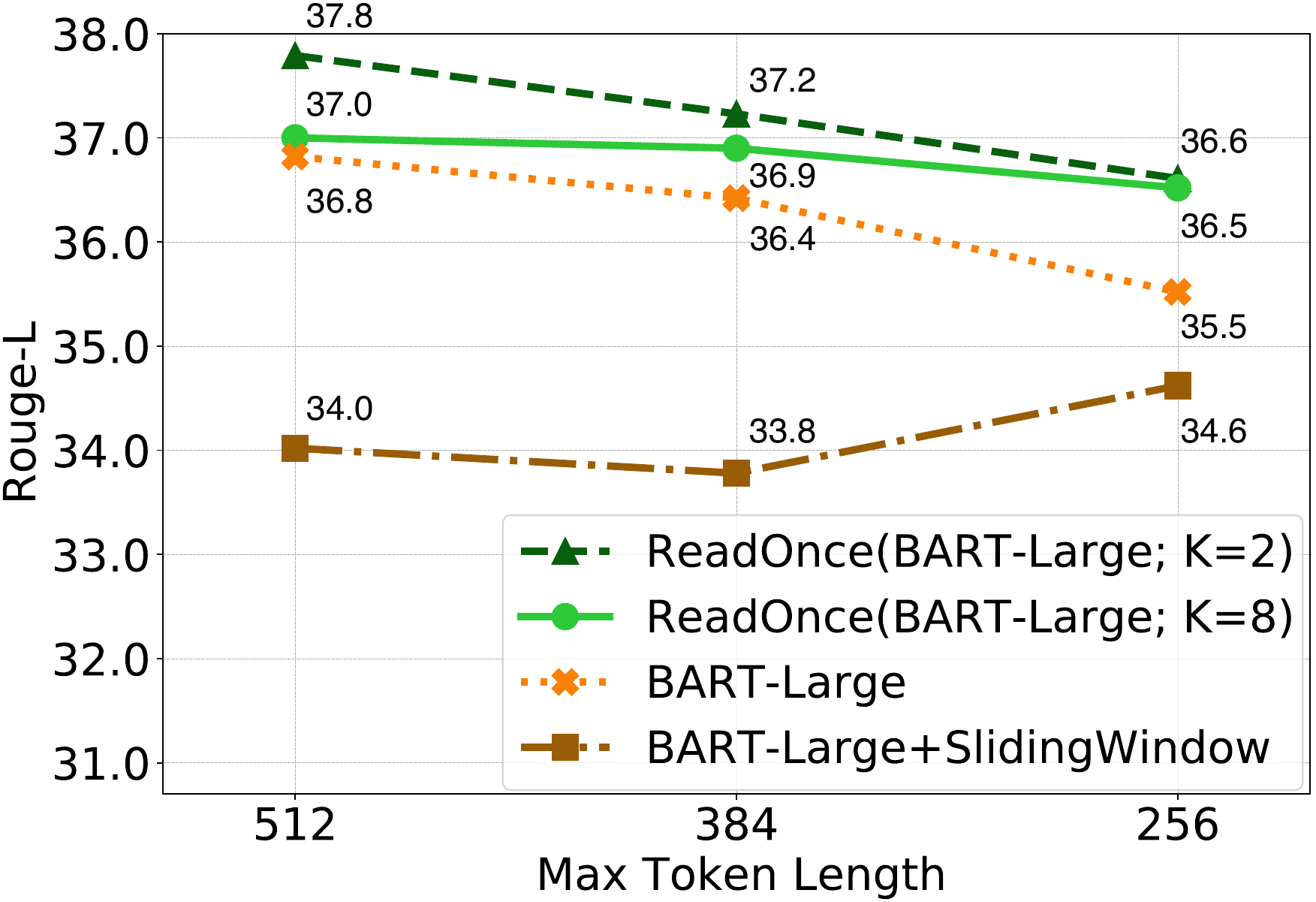}
    \caption{Accuracy of models under different maximum window length assumptions on PubMed dataset. \sysname stay substantially more accurate as the maximum window length decreases.}
    \label{fig:long_doc_pubmed}
\end{figure}

ROUGE-L scores of these models are depicted in Figure~\ref{fig:long_doc_pubmed}. As we reduce T for the underlying transformer model from 512 to 256, the score of the baseline \bart model drops to 35.5 ROUGE-L. When used with the sliding window technique, the performance is even worse, likely due to the naive aggregation of the summaries. Our approach, on the other hand, concatenates document representations, allowing the downstream model to build a coherent summary. We see the ROUGE-L score only drops to 36.6 when K=2 (with model capacity dropping from 1024 to 512 tokens) and a much smaller drop from 37.0 to 36.5 when K=8 (with model capacity dropping from 3520 to 1472 tokens). This simulation shows that concatenating \sysenc is a simple yet effective way to increase the capacity of existing models.

\section{Conclusion}

This work introduced \sysname, a novel approach for using large scale transformer-based language models to both build and consume reusable document representations. Akin to humans' ability to read a document and extract useful information without knowing the end-use, \sysenc are compact, information-capturing document representations that can be pre-computed once, in a task- and example-independent fashion.

Our results on extractive QA, summarization, and abstractive QA tasks demonstrate that using \sysenc, in lieu of re-reading document text in the context of every example, results in substantially faster training and inference, at a modest cost in accuracy. The \sysshortname framework also offers an easy way to control the trade off between speed and accuracy (via the compression ratio parameter), and enables the use of standard transformer architectures on long documents beyond the model's token limit.

Identifying the ideal compact document representations in our controlled setting opens up the possibility of efficient open-domain QA, where models retrieve and reason directly over these representations. We leave an exploration of the training of the retrieval function, often with only answer supervision and ideally in an end-to-end setting, to future work.

\subsection*{Acknowledgements}
We thank Dirk Groeneveld for providing the output of the Quark system for \hotpot and the Beaker team for their support with the  experiments.

\bibliographystyle{acl_natbib}
\bibliography{readonce}

\clearpage
\appendix
\section{Model Details}
\subsection{The Detailed Architecture of Modified Transformer Block Attention}
\label{app:plotmachines}
In this variation of \RepTextNoc, the modified self-attention block uses two separate attention modules for both of these input types and averages  the vectors. Specifically, let $\mathbf{H}_{\mathrm{enc}}^{L}$ be the matrix of hidden states generated from the L\textsuperscript{th} layer of a standard transformer:
\begin{equation}
    \mathbf{H}_{\mathrm{enc}}^{L} = \mathrm{Attn}(\mathbf{H}_{\mathrm{enc}}^{L-1}, \mathbf{H}_{\mathrm{enc}}^{L-1}, \mathbf{H}_{\mathrm{enc}}^{L-1})
\end{equation}
where $\mathrm{Attn}(\mathbf{Q}, \mathbf{K}, \mathbf{V})$ is the attention module used in the transformer that takes $\mathbf{Q}, \mathbf{K}, \mathbf{V}$ as the query, key, and value matrices. To take extra \sysenc as an input, we instead compute $\mathbf{H}_{\mathrm{enc}}^{L}$ as:
\begin{equation}
\label{eq:plotmachine}
\begin{split}
    \mathbf{H}_{\mathrm{enc}}^{L} = &(\mathrm{Attn}(\mathbf{H}_{\mathrm{enc}}^{L-1}, \mathbf{H}_{\mathrm{enc}}^{L-1}, \mathbf{H}_{\mathrm{enc}}^{L-1}) + \\
    &\mathrm{Attn}^{'}(\mathbf{H}_{\mathrm{enc}}^{L-1}, \mathbf{R}, \mathbf{R})) / 2
\end{split}
\end{equation}
where $\mathrm{Attn}^{'}(\cdot)$ is a separate attention module to include the \sysenc $\mathbf{R}$ in our \RepTextNoc, whose weights are initialized by the corresponding weights in $\mathrm{Attn}(\cdot)$ to speed up the training. For the decoder in the \RepTextNoc, we also compute the hidden states of each layer as per Eqn. \ref{eq:plotmachine} so that the model can attend over the extracted document information during the decoding process too.

\section{Dataset Details}
\label{app:datasets}

    (1) \textbf{\hotpot}~\cite{hotpotqa} is a multi-hop reasoning dataset that requires models to aggregate information from two paragraphs to produce the answer (a span from the input paragraphs). We focus on their \textit{distractor} setting where they additionally provide models with 8 distractor paragraphs. For efficiency, we use the output of the Quark system~\cite{groeneveld2020simple} which selects up to 512 tokens (including the question) from the input paragraphs. We use the answer EM and F1 scores as the metrics.
    
    (2) \textbf{XSUM}~\cite{Narayan2018DontGM} is an abstractive News summarization dataset that requires models to generate summaries that go beyond simply extracting key sentences. We use the Rouge-L Summ.\ score\footnote{\url{https://github.com/google-research/google-research/tree/master/rouge}} commonly used for summarization datasets, which computes the union-LCS  of the longest common subsequences (LCS) between each pair of reference and hypothesis sentences. In contrast, the standard Rouge-L score computes LCS between the reference and hypothesis, treating both of them as one sentence. 
    
    (3) \textbf{NarrativeQA}~\cite{Kocisk2018NarrativeQA} is an abstractive QA dataset where answers may not be extractive spans in the input document. Models would need to understand the content of the document to generate such answers. We use the same Rouge-L Summ.\ score as for the summarization task.\footnote{In our experiments, we did not notice any substantial difference between the simple Rouge-L metric and this summarization-based metric.}

    (4) \textbf{PubMed}~\cite{Cohan2018ADA} is an abstractive long-document summarization dataset specifically focusing on the scientific publications. The large number of tokens in each document makes it hard for standard Pretrained Transformers to deal with. We use the same Rouge-L Summ.\ score as for XSUM.
    
\section{Experiment Setting for DistilBART}
\label{appendix:distillation}
We follow \citet{distilbart} to obtain our DistilBART model used in \S\ref{sec:exp-downstream-tasks}. Specifically, we first create a student model with 12-layer encoder and 6-layer decoder from BART-Large$_\text{{\squad+\unsup}}$ using  the ``Shrink and Fine-Tune" distillation described in \citet{distilbart}, which has been shown to be effective for BART model on summarization tasks. We then further finetune the student model on SQuAD+UQA, and exploit the resulting model as our DistilBART. 
    

\subsection{Setup for the Pubmed Dataset in the Long-document Experiment}
\label{appendix:pubmed-setup}
 We follow \citet{Cohan2018ADA} and only include 4 sections from document. After the truncation, the average number of tokens in the documents in this dataset is 2270, with 90\% of the documents being under 4253 tokens. To include the information from each section, we evenly distribute the length budget T across the sections. This would mean, for the baseline BART seq2seq model, each section is first truncated to T/4 tokens, then the 4 sections are concatenated as the input. As for \sysname, we first compute representations for K/2 chunks of each section with length T/2 tokens, then aggregate them as the final \sysenc for the input document. In this case, even when K equals 2, we are allowed to include one chunk from each section without exceeding the length limit. For the BART + SlidingWindow baseline, we concatenate summaries from 16 chunks of length T with 4 chunks from each section.

\subsection{Handling Long Documents: Narr.QA}
\label{appendix:longdoc_nqa}
\begin{figure}[ht]
    \centering
    \includegraphics[width=0.48\textwidth]{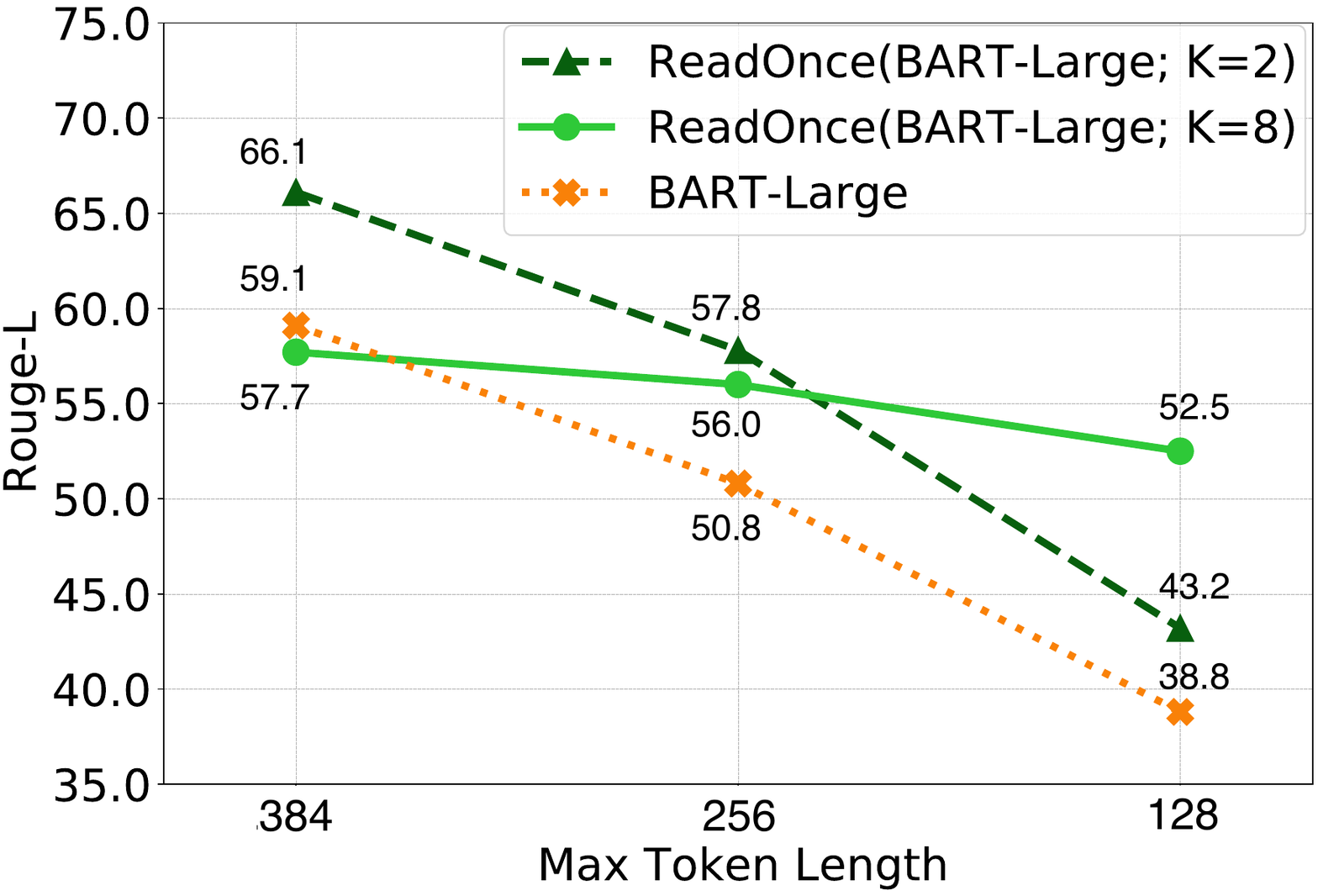}
    \caption{Accuracy of models under different maximum window length assumptions on NarrativeQA dataset. \sysname stay substantially more accurate as the maximum window length decreases.}
    \label{fig:long_doc_nqa}
\end{figure}

Aside from Pubmed, we also evaluate the ability of \sysname to handle long documents on the NarrativeQA dataset. The average number of tokens in the documents in this dataset is 668.6, with 90\% of the documents being under 981 tokens. The results are depicted in Figure~\ref{fig:long_doc_nqa}.

As we reduce T for the underlying transformer model from 384 to 128, the score of the baseline \bart model drops significantly from 59.1 to 38.8. With K=2, our model consistently outperforms the baseline model but exhibits a similar drop in score as the maximum token limit of this model with T=128 is still only 208 tokens (as per the equation above). On the other hand, with K=8 and T=128, we can handle documents up to 688 tokens, thereby requiring no truncation on 50\% of the examples even in this extreme scenario. As a result, we only see a 5.2 point drop in score as T is decreased from 384 to 128. These simulations provide strong evidence of the ability of \sysname to handle long documents more effectively than standard transformer based models.

\subsection{Read-Once Efficiency Gains}
\label{subsec:efficiency}

Let $C$ denote context length (as \#tokens), $Q$ the question length, $R$ the repetition factor (i.e., \#questions per context), and $K$ the \sysshortname compression factor. For an input of $N$ tokens, we treat the computational cost of the encoder (forward or backward pass) as $T_e N^2$ and that of the decoder as $T_d N^2$, where $T_e, T_d$ capture the complexity of the encoder and decoder model respectively and $N^2$ captures the self-attention over the input context. We ignore the self-attention over the decoded text as it is unchanged across models.

For any baseline seq2seq model, the computational cost of processing an example (context+question)  can be captured by $(T_e + T_d) (C+Q)^2$.
For \sysname, the cost of computing a context's representation, amortized over the $R$ questions that share this context, is $T_e \frac{C^2}{R}$. Once the compressed context representation is appended to the question encoder at layer $L$ (out of 12), the rest of the encoder computation costs $T_e \cdot \left( \frac{L}{12} Q^2 + \left(1 - \frac{L}{12}\right) \left(\frac{C}{K} + Q\right)^2 \right)$. The decoder's computation cost is $T_d \cdot \left(\frac{C}{K} + Q\right)^2$.

When $L=6$ and $Q \ll \frac{C}{K}$, the net computational cost (without caching) simplifies to $\frac{T_e C^2}{R} + \frac{T_e C^2}{2 K^2} +  \frac{T_d C^2}{K^2}$. Assuming $T_e \approx T_d = T$, this equals $T C^2 \left(\frac{1}{R} + \frac{3}{2K^2}\right)$. In contrast, the baseline model's cost simplifies to $2 T C^2$. The efficiency gain of \sysname over the baseline encoder-decoder model is therefore roughly $\frac{1}{2}\left(1/R + 3/2K^2\right)$.

Additionally, when we use these representations for multiple training runs, inferences, downstream tasks, etc., the cost of computing the fixed representations is basically amortized to a constant term. As a result, over multiple runs, using \sysenc now reduces the cost of the building and using models to just $T C^2 \frac{3}{2K^2}$. So using these cached representations amortized over multiple epochs/runs, improves the efficiency gains further to $3/4K^2$.



\end{document}